\documentclass[review,10pt]{elsarticle} %Elsevier document class
\usepackage[a4paper,margin=2.5cm]{geometry} %Margin specifications
\usepackage[utf8]{inputenc} %Encoding for some characters
\usepackage[english]{babel}
\usepackage{caption}
\usepackage{subcaption}
\usepackage{array}
\usepackage{lineno} %Show line numbers
\usepackage[colorlinks=true]{hyperref} %Color hyperlinks
\usepackage{float}
\usepackage{multirow}
\usepackage{xurl}
\usepackage{tabularx,ragged2e,booktabs,caption}
\usepackage{amsmath}

\begin{document}
\begin{frontmatter}
	
	\title{Offshore Wind Plant Instance Segmentation Using Sentinel-1 Time Series, GIS, and Semantic Segmentation Models}
	
	%% Authors and affiliations:
	\author[icc]{Osmar Luiz Ferreira de Carvalho}
	\ead{osmarcarvalho@ieee.org}
	\author[unb]{Osmar Abílio de Carvalho Júnior \corref{corresp}}
	\author[unb]{Anesmar Olino de Albuquerque}
	\ead{anesmar@ieee.org}
	\author[icc]{Daniel Guerreiro e Silva}
	\ead{danielgs@unb.br}

	\affiliation[icc]{
		organization={University of Brasilia},
		department={Department of Electrical Engineering,},
		addressline={Campus Universitario Darcy Ribeiro}, 
		city={Brasilia},
		postcode={70910-900}, 
		state={Federal District},
		country={Brazil}
	}
	\affiliation[unb]{
		organization={University of Brasilia},
		department={Departament of Geography,},
		addressline={Campus Universitario Darcy Ribeiro}, 
		city={Brasilia},
		postcode={70910-900}, 
		state={Federal District},
		country={Brazil}
	}

	\cortext[corresp]{Corresponding author: osmarjr@unb.br}

\begin{abstract}
Offshore wind farms represent a renewable energy source with a significant global growth trend, and their monitoring is strategic for territorial and environmental planning. This study's primary objective is to detect offshore wind plants at an instance level using semantic segmentation models and Sentinel-1 time series. The secondary objectives are: (a) to develop a database consisting of labeled data and S-1 time series; (b) to compare the performance of five deep semantic segmentation architectures (U-Net, U-Net++, Feature Pyramid Network - FPN, DeepLabv3+, and LinkNet); (c) develop a novel augmentation strategy that shuffles the positions of the images within the time series; (d) investigate different dimensions of time series intervals (1, 5, 10, and 15 images); and (e) evaluate the semantic-to-instance conversion procedure. LinkNet was the top-performing model, followed by U-Net++ and U-Net, while FPN and DeepLabv3+ presented the worst results. The evaluation of semantic segmentation models reveals enhanced Intersection over Union (IoU) (25\%) and F-score metrics (18\%) with the augmentation of time series images. The study showcases the augmentation strategy's capability to mitigate biases and precisely detect invariant targets. Furthermore, the conversion from semantic to instance segmentation demonstrates its efficacy in accurately isolating individual instances within classified regions—simplifying training data and reducing annotation effort and complexity.
\end{abstract}

	
\end{frontmatter}

\pagebreak
%\linenumbers
\section{Introduction}
Offshore wind energy (OWE) has emerged as a significant and rapidly expanding contributor to the global power generation portfolio, driven by a compelling blend of environmental sustainability, technological advancements, and policy support. Unlike their onshore counterparts, offshore wind plants, strategically located in oceanic or sea regions, favor larger-scale constructions, take advantage of the more consistent and stronger winds available in these open spaces to generate renewable electricity, and reduce noise pollution by being further away from populated areas \citep{sun2012current, zheng2016overview}. Furthermore, the advent of floating wind turbines will expand offshore wind energy beyond shallow waters, not relying on the ocean floor and conventional gravity-based concrete at depths and substantially increasing the potential of the world's oceans \citep{bashetty2021review, mcmorland2022operation}. These advances also provide a promising renewable energy source solution to the inherent spatial constraints of densely populated regions or those with limited land availability.

Therefore, OWE is at a significant inflection point in its evolution due to technological advances, economies of scale, and increased political support, making it a more cost-competitive energy source and reducing greenhouse gas emissions \citep{hasager2015offshore}. According to the World Forum Offshore Wind, global offshore wind capacity in operation surpassed 57 GW at the end of 2022, representing an additional 9.4 GW during 2022 \citep{WFO2023}. China has the highest installed capacity, with 26.6 GW, followed by the United Kingdom (UK) in second place, reaching 13.6 GW. In 2022, China led the global offshore wind capacity under construction, with a total capacity of 3.4 GW in Chinese waters. The UK follows closely in second place with 2.8GW, which includes notable projects like Dogger Bank A (1.2GW) and Seagreen (1.1GW), both with a capacity in the gigawatt range (WFO, 2023). The International Energy Agency (IEA) presented two scenarios for global offshore wind capacity growth by 2040: the State Policy Scenario, driven by policy targets and falling technology costs, and the Sustainable Development Scenario, which includes incentives to accelerate decarbonization efforts in the electricity sector \citep{IEA2019}. Under the Sustainable Development Scenario, global offshore wind capacity will reach approximately 560 GW by 2040, representing a 65\% increase compared to the State Policy Scenario. 

However, exponential growth projections on a global scale for offshore wind energy infrastructures involve several interests and conflicts with legal, socio-economic, and ecological implications \citep{glasson2022local}. Despite OWE's clear contributions to addressing the global emergence of carbon-neutral energy, this technology has limitations and encompasses wide-ranging challenges. As the OWE intensifies and expands, the tendency is to increase conflicts with nearby coastal communities with interests contrary to the use of marine environments, such as aquaculture, tourism, navigation, and commercial fishing \citep{virtanen2022balancing}. OWE negatively impacts diverse marine life populations due to collision mortality, disruption of migration corridors, barrier effects, acoustic and electromagnetic disturbances, ecological function changes, habitat alteration, and effects on the benthic ecosystem \citep{abramic2022environmental, bergstrom2014effects, lane2020vulnerability, li2023offshore}. Besides, OWE growth between 2020 and 2040 will require significant quantities of raw materials, including 8.2-14.6 million tons (Mt) of iron, 129-235 Mt of steel, 3.8-25.9 Mt of concrete, 0.5-1.0 Mt of copper, 0.3-0.5 Mt of aluminum, as well as substantial amounts of rare earth elements \citep{li2022future}. 

Monitoring the growth and analyzing the spatial distribution of offshore wind farms provide valuable information for formulating policies for the efficient use of available marine space, minimizing conflicts with other activities, such as maritime routes, fishing grounds, and environmentally sensitive areas. This spatiotemporal information helps to alert and plan a balance in generating renewable energy and the sustainable use of marine resources. However, the on-site monitoring of these large-scale installations, spread over vast geographic expanses, is logistically complex and economically prohibitive \citep{leung2012wind}. In this context, remote sensing technology presents a huge potential to provide high-quality data for automatically monitoring offshore wind plants, overcoming the logistical complexities of manual inspections. 

Coupled with remote sensing images, recent advances in deep learning models achieve state-of-the-art object detection and segmentation in digital image processing \citep{liu2020deep, ma2019deep}. Therefore, deep-learning methods have promoted a revolution in computer vision and remote sensing, mainly using Convolutional Neural Networks (CNNs) architectures, obtaining accuracy results that surpass other approaches \citep{li2022future}. These sophisticated models can learn to distinguish between different landscape features, differentiate between wind farms and other structures, and perform these tasks under varying conditions, making them a promising solution to the detection challenges of offshore wind farms. Previous studies aimed at detecting wind farms using remote sensing and deep learning predominantly focused on onshore areas. These onshore wind plant studies considered different associations of optical images and CNN architectures: (a) Gaofen-2 images and U-Net \citep{han2018targets}; (b) aerial photographs and comparative analysis between U-Net and LinkNet \citep{manso2020optimizing}; (c) aerial photographs and multiclass reconnaissance network \citep{manso2021first}; (d) aerial photographs and R-CNN Mask \citep{schulz2021deteektor}; and (e) China-Brazil Earth Resources Satellite (CBERS) 4A scenes and CNN architecture comparative analysis (U-Net, U-Net++, LinkNet DeepLabv3+, Feature Pyramid Network - FPN) \citep{de2023data}.

In the context of mapping offshore wind farms, the first studies used traditional techniques considering different methodological approaches, image data, and study areas: (a) cloud-based geoprocessing algorithm using synthetic aperture radar (SAR) and the Google Earth Engine (GEE) to detect offshore infrastructure, including oil platforms and wind turbines in the Gulf of Mexico \citep{wong2019automating}; (b) visual saliency detection (VSD) algorithm based on optical satellite images from various sources (Landsat-5 Thematic Mapper, Landsat-7 Thematic Mapper Plus, Landsat-8 Operational Land Imager, and Sentinel-2 Multi-Spectral Instrument) in the North Sea and surrounding waters \citep{xu2020proliferation}; (c) percentile-based reduction algorithm using Sentinel-1 SAR time-series images, constructing a dataset containing over 6,900 turbines in 14 coastal nations \citep{zhang2021global}; and (d) random forest (RF) method using Sentinel-1 data on the GEE in the Yellow Sea of China and the North Sea of Europe \citep{xu2022dynamic}.

Researcher Thorsten Hoeser led the published research involving offshore wind plant detection based on deep learning techniques, all using Sentinel-1 data and object detection methods, considering the following purposes: (a) development of the DeepOWT dataset with space-time information on offshore wind energy infrastructure on a global scale \citep{hoeser2022deepowt}; (b) the proposition of SyntEO, a system that generates large datasets for deep learning, merging existing and synthetic data through the incorporation of specialized and structured knowledge \citep{hoeser2022synteo}; and (c) estimate of the global capacity of offshore wind turbines based on the analysis of spatiotemporal patterns of offshore wind turbines, revealing substantial growth of the sector due to the European Union, China, and the United Kingdom \citep{hoeser2022global}.

Despite significant advances in the study of offshore wind farms using object detection techniques, there is an open opportunity to employ semantic and instance segmentation methods. Object detection aims to identify the location of various objects in an image and establish their bounding boxes. In opposition, semantic and instance segmentation aim at a pixel-level classification, assigning a class and instance label. This paper provides methodological and practical novelties, in which the most significant contributions are:

\begin{enumerate}
\item Comparative analysis of semantic segmentation methods: This research performs and compares semantic segmentation algorithms in the context of offshore wind farms, unlike using object detection techniques.

\item Publicly available dataset: Construction of a robust dataset available for public use, comprising more than 5,000 patches, each with dimensions of 128 x 128 x time sequences (1, 5, 10, and 15 images). This unique dataset serves as a valuable resource for future research and development of algorithms in detecting and segmenting offshore wind farms using time series as training data.

\item Simplified instance segmentation: By employing semantic segmentation, a more straightforward deep learning method, we streamline the process to achieve instance-level segmentation. Coupling this approach with Geographic Information System (GIS) tools allows us to detect and segment wind turbines more effectively, setting a precedent for future research and applications.
\end{enumerate}

%\pagebreak
\section{Materials and methods}
\subsection{Dataset Construction}

The study area focused on the United Kingdom (UK) (Figure 1), a global pioneer in offshore wind technology \citep{higgins2014evolution}. With the second largest offshore wind capacity worldwide, the UK presents a unique and invaluable landscape for understanding the dynamics and complexities of these large-scale renewable energy installations, including significant spatial, seasonal, and regional variations \citep{potisomporn2022spatial}. Therefore, this study carefully selected ten regions distributed across the UK, where each location has distinct environmental conditions, wind farm sizes, and technology configurations, offering a comprehensive and diverse set of scenarios for analysis. These regions host substantial offshore wind farms, contributing significantly to the UK's renewable energy portfolio. Given their geographical, technological, and environmental attributes, the variety in data from these sites provides a rich resource for understanding the potential and challenges of automated offshore wind plant detection using remote sensing data, besides evaluating the robustness, accuracy, and scalability of semantic segmentation models learning under a range of conditions. Furthermore, we aim to identify critical factors and patterns that may influence the effectiveness of our proposed methods.

\begin{figure}[!h]
	\centering %
	\scriptsize %
	\includegraphics[width=\columnwidth]{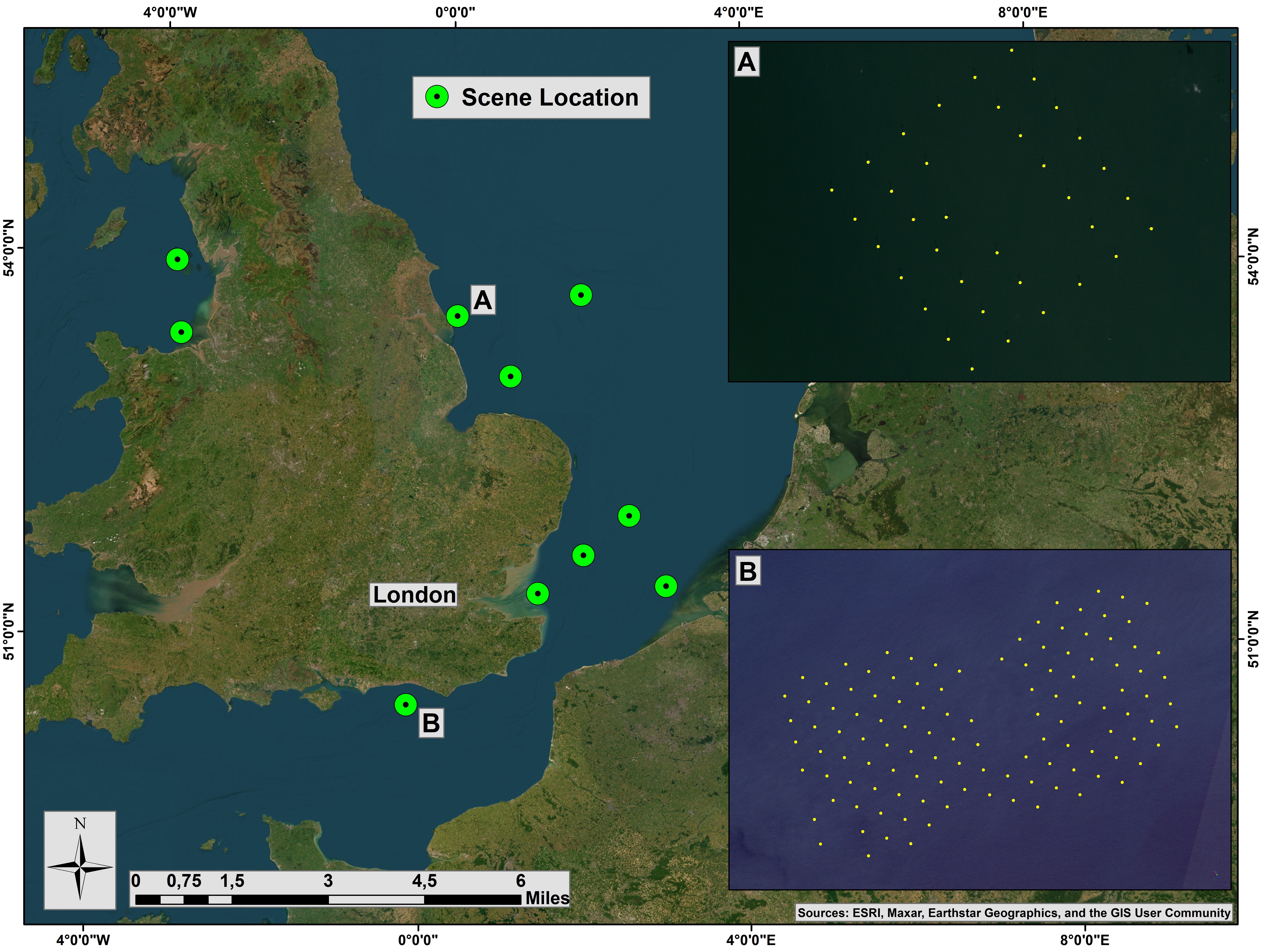}
	\caption{Location of the sites containing offshore wind plants.}
	\label{fig:fig1}
\end{figure}

In contrast to the challenge of detecting onshore wind farms using SAR images – due to the complexities arising from the lack of evident distinction in the terrestrial environment – offshore wind farms have larger dimensions, and over the vast ocean produces a high-backscatter pattern that facilitates their detection. Therefore, this research acquired images from the SAR Sentinel-1 mission developed by the European Space Agency (ESA), which comprises a constellation of two polar-orbiting satellites, Sentinel-1A and Sentinel-1B. These sensors operate in the 5.4 GHz C-band frequency and capture synthetic aperture radar (SAR) images in VV (vertical transmission, vertical reception) and VH (vertical transmission, horizontal reception) polarizations, with free access on the Copernicus Open Access Hub. The data specifications were Interferometric Wide Swath (IW) images with VV polarization in Level-1 Ground Range Detected (GRD) format. Using VV polarization is due to its lower noise influence and ability to discern better the target characteristics in the sea \citep{de2022deep, zhang2021global}. The analysis encompassed a time series of up to 15 SAR images captured in 2022 and 2023. The image pre-processing used the Sentinel Application Platform (SNAP) software, including the following steps \citep{filipponi2019sentinel}: (a) apply orbit file, (b) thermal noise removal, (c) calibration to convert the digital pixel values to radiometrically calibrated SAR backscatter, (d) speckle filtering using the Lee Sigma filter, (e) range-Doppler terrain correction utilizing the Global Earth Topography and Sea Surface Elevation at 30 arc-second resolution data, and (f) conversion of backscattering values to decibels (dB).

\subsection{Ground Truth Generation}
In the ground truth annotation, the inconsistent sunlight reflection of the metallic structure of the wind farms in each image was a significant challenge, making the delineation process complex (Figure 2). Despite the noise, many pixels, particularly those in the wind farm core, exhibited consistently high brightness across time series. Taking advantage of this consistency, we create a mask by averaging all images in the time series (composite image) and classifying high-backscatter elements that exceed 0 in the VV band as positive. This approach resembles the proposal developed by \cite{zhang2021global}. The data refinement used vector format (shapefile) in the ArcGIS software, enabling manual noise elimination. In this annotation task, the optical images served as a support to differentiate the target from other elements, such as oil platforms or boats.

Offshore wind farm detection using time-series data can overcome difficulties in detecting wind farms, increasing the chances of obtaining clear and discernible images of wind farms and removing floating or temporarily mobile objects. Therefore, time series data strengthens the learning process of deep architectures to mitigate problems and ensure reliable results, allowing us to apprehend the invariant behavior of wind farms face to other moving features or noise and identify their morphological variations due to backscattering patterns.

\begin{figure}[!h]
	\centering %
	\scriptsize %
	\includegraphics[width=\columnwidth]{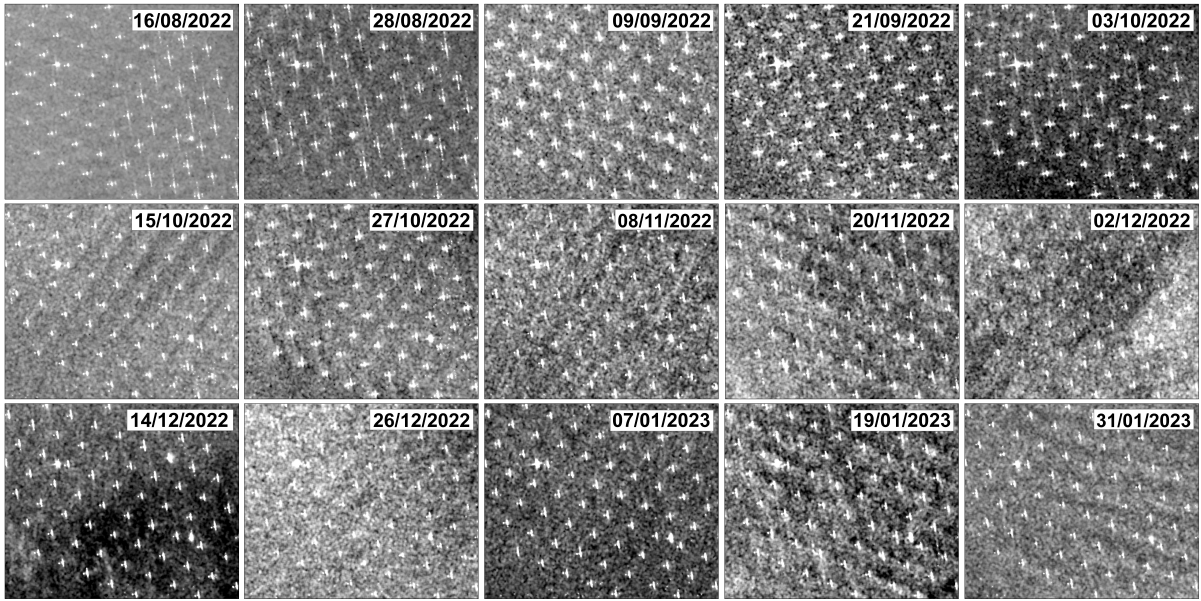}
	\caption{Time series of 15 images showing the offshore wind plants.}
	\label{fig:fig2}
\end{figure}

\subsection{Semantic Segmentation Dataset Generation}
Sample preparation containing the image and its corresponding ground truth mask is crucial to train semantic segmentation models. This process begins by breaking down each full-size image into smaller, fixed-size pieces, known as patches. The spatial distribution and cutout of the patches used the methodology developed by \cite{de2022panoptic}, which employs user-supplied central points (shapefile format) to delineate the frame. Strategic choice of patches assures an efficient data selection rather than systematic clipping through the sliding window. This study selected a patch size of 128x128 pixels, yielding 4947 patches and encompassing 2910 distinct offshore wind plants. The patch count outweighs the number of wind farms, aiming to capture wind plants under various scenarios to bolster the robustness of the dataset. Additionally, the incorporation of background element patches without the presence of wind farms enriched the overall data.

The training, validation, and testing split considered distinct sets to avoid overlap. Therefore, six of the ten chosen areas were allocated for training, two for validation, and the other two for testing. Table 1 lists the characteristics of each location, including its assigned data set, the number of patches, and the count of wind plants. Table 2 lists the synthesized characteristics of the dataset. This careful distribution ensures balanced exposure to various conditions and scenarios, enhancing the dataset's overall efficacy and utility.

\begin{table}[!h]
\centering
\setlength\extrarowheight{-4pt}
\caption{Detailed description of the dataset, including the number of instances, patches, location of data collection, data partitioning into training, validation, testing subsets, and the time interval from the first and last image in the time series.}
\begin{tabular}{l|l|l|l|l}
\hline
Location & \# of wind plants & \# of patches & Split & Time interval \\
\hline
3°36'9,645"W 54°3'2,492"N     & 374 & 652 & Train & 14/08/22 to 29/01/23 \\
3°26'37,919"W 53°29'28,928"N  & 269 & 565 & Train & 14/08/22 to 29/01/23 \\
0°9'3,647"E 53°48'23,561"N    & 70  & 131 & Train & 16/08/22 to 31/01/23 \\
0°54'38,874"E 53°16'50,015"N  & 474 & 611 & Val   & 16/08/22 to 31/01/23 \\
1°46'53,292"E 53°55'11,708"N  & 308 & 402 & Test  & 16/08/22 to 31/01/23 \\
1°22'3,046"E 51°33'52,405"N   & 387 & 903 & Train & 16/08/22 to 31/01/23 \\
1°57'20,788"E 51°51'40,077"N  & 220 & 557 & Test  & 16/08/22 to 31/01/23 \\
2°31'22,737"E 52°12'26,383"N  & 96  & 144 & Train & 16/08/22 to 31/01/23 \\
2°58'19,339"E 51°39'41,574"N  & 592 & 708 & Val   & 16/08/22 to 31/01/23 \\
0°16'34,541"W 50°40'11,965"N  & 120 & 274 & Test  & 09/08/22 to 17/02/23 \\
\hline
Total & 2910 & 4947 & - & - \\
\hline
\end{tabular}
\label{tab:tab1}
\end{table}

\begin{table}[!h]
\centering
\setlength\extrarowheight{-4pt}
\caption{Synthesized table with the number of unique instances and number of patches for each set of data.}
\begin{tabular}{l|l|l}
\hline
Set & \# of wind plants & \# of patches \\
\hline
Train & 1,196 (41.1\%) & 2,395 (48.41\%) \\
Val   & 1,066 (36.63\%) & 1,319 (26.27\%)\\
Test  & 648 (22.27\%)  & 1,233 (24.92\%) \\
\hline
\end{tabular}
\label{tab:tab2}
\end{table}

%\pagebreak
\subsection{Deep Learning approach}
\subsubsection{Deep learning models}
This study compared a broad range of deep learning architectures known for their proficiency in semantic segmentation tasks. Our selection included Deeplabv3+ \citep{chen2018encoder}, U-Net \citep{ronneberger2015u}, U-Net++ \citep{zhou2018unet++}, Feature Pyramid Network (FPN) \citep{lin2017feature}, and LinkNet \citep{chaurasia2017linknet}:

\begin{itemize}
    \item Deeplabv3+: This architecture introduces an encoder-decoder structure to refine the segmentation results, especially along object boundaries. At its core, it uses atrous convolution with atrous spatial pyramid pooling (ASPP) to robustly capture multi-scale context.
    
    \item U-Net: This architecture has a U-shaped model. The encoder captures context by down-sampling the input, while the decoder enables precise localization through a symmetric up-sampling path. A critical feature of U-Net is the presence of skip connections between encoder and decoder layers at corresponding resolutions, facilitating the fusion of low-level feature maps with high-level ones, thereby preserving fine-grained details during up-sampling.
    
    \item U-Net++: This architecture leverages the U-Net by introducing nested and dense skip architecture. The nested structure facilitates feature reuse among different paths, while the dense skip connections provide more direct paths for gradients during back-propagation, improving feature propagation and model performance. This robust design reduces the semantic gap between the encoder's and decoder's feature maps, leading to better feature fusion and enhanced segmentation performance.

    \item FPN: This architecture is a robust object detection framework that tackles multi-scale object detection challenges, which was also adapted for the semantic segmentation task. It constructs a multi-scale, pyramidal hierarchy of feature maps by incorporating a top-down architecture and lateral connections into a conventional CNN. The bottom-up pathway extracts a multi-scale feature hierarchy from a single-scale input. The top-down pathway enhances the resolution of features, utilizing both the feed-forward feature and its corresponding feature from the bottom-up pathway. Lateral connections merge feature maps from both pathways. 

    \item LinkNet: This model uses a simplified encoder-decoder structure for tasks such as image segmentation. Its power lies in proficiently restoring intricate image details from high-level features. This balancing act between computational efficiency and segmentation precision makes it an asset for various vision tasks. The encoder extracts features, and the decoder, paralleling the encoder's architecture, reconstructs spatial information, preserving both the context and details of the image.
\end{itemize}

Moreover, the EfficientNet-B7 was the backbone of all models. EfficientNet \citep{tan2019efficientnet} is a state-of-the-art CNN offering superior accuracy and computational efficiency. What distinguishes it is its compound scaling method, which uniformly scales all dimensions of the network's depth (number of layers), width (number of channels), and resolution (input image size) while maintaining a balanced network structure. The foundational EfficientNet-B0 model uses AutoML and reinforcement learning as the baseline network. The subsequent EfficientNet variants (B1-B7) are improved versions of this baseline model, with each version delivering progressively higher accuracy while adhering to the compound scaling rule.

The hyperparameters were the same for all models to ensure a consistent comparison: (1) Adam optimizer, (2) learning rate of 0.001, (3) batch size of 20, and (4) 150 epochs. Model selection considered the cross-entropy loss in the validation set, saving the parameters that led to the smallest validation loss for each architecture for further statistical analysis.

\subsubsection{Novel data augmentation strategy}
While time series typically depend on the data frames' sequential order, as for agriculture and phenology studies, our research focuses on a time-invariant target. In this context, the image sequence within the time series is irrelevant and could introduce bias. If a model is invariably trained on an identical sequence of images, there is a risk of inadvertently learning specific patterns from that sequence. This arrangement could create problems concerning its capacity to generalize and perform accurately on diverse datasets.

We propose a novel data augmentation technique tailored for semantic segmentation tasks to circumvent this potential pitfall. This innovative strategy involves randomly shuffling the order of the images in the time series during the model's training process. This measure ensures that the model does not unduly familiarize itself with any specific sequence and consequently improves its generalization ability. Our approach, thus, paves the way for a more robust and reliable model that can effectively handle diverse time-invariant datasets. A similar approach has been made, for instance segmentation of center pivots using optical imagery, in which the main objective was to avoid the interference of clouds in the model \citep{de2021dealing}.

\subsubsection{Experiments}
The experiment manipulated the time-series interval, analyzing datasets with 1, 5, 10, and 15 images. Even with the preservation of training, validation, and testing sampling sites, the analysis established different time-series periods. The proposed augmentation approach randomly flips the order of images in each dataset (training, validation, and test). However, to maintain consistency in comparing models’ performance, we performed a single shuffle in the validation and test samples, preserving their ordering across all evaluated models. Furthermore, the efficiency analysis of the new augmentation strategy compared the best-performing model with and without its implementation in the time series.

\subsubsection{Sliding Windows approach}
Practical applications of semantic segmentation for remote sensing data often rely on a sliding window approach. This method facilitates the comprehensive classification of images, typically much larger than the patches used for model training. The sliding window approach involves a systematic shift of a window, equivalent to the size of the training patches, across the image in both x and y dimensions, using a predetermined stride value. The largest permissible stride value corresponds to the window's dimension. Stride values less than this result in overlapping pixels. These overlaps enhance the resulting classification by mitigating the discontinuities between adjacent frames, creating a smoother and more coherent segmentation output \citep{costa2021remote}. This technique ensures accurate and detailed semantic segmentation across large remote-sensing datasets.

\subsubsection{Metrics}
This research considered two types of accuracy assessment metrics: per-pixel and per-object approaches. Per-pixel metrics are valuable to understand how many prediction pixels matched the reference pixels in patches from the test set. In this context, the accuracy analysis used the following metrics: overall accuracy, precision, recall, f-score, and Intersection over Union (IoU) (Table 3).

\begin{table}[h!]
\centering
\begin{tabular}{|l|l|}
\hline
\textbf{Metric}       & \textbf{Formula}                           \\ \hline
Overall Accuracy      & \((TP + TN) / (TP + TN + FP + FN)\)        \\ \hline
Precision             & \(TP / (TP + FP)\)                         \\ \hline
Recall                & \(TP / (TP + FN)\)                         \\ \hline
f-score               & \(2 \times ((\text{Precision} \times \text{Recall}) / (\text{Precision} + \text{Recall}))\) \\ \hline
IoU                   & \(TP / (TP + FP + FN)\)                    \\ \hline
\end{tabular}
\caption{Mathematical formulation for overall Accuracy (OA), Precision, Recall, F-score, and Intersection over Union (IoU), considering True Positives (TP), True Negative (TN), False Positives (FP), and False Negatives (FN).}
\label{tab:tab3}
\end{table}

The per-polygon metrics quantify the number of objects classified correctly, establishing a correspondence between a segmentation's mapped and reference polygons. This approach minimizes the effects of errors in target contours and focuses on matching object counts. Per-polygon performance analysis quantified true positives (TP), false positives (FP), and false negatives (FN) and calculated the following metrics: Overall Quality (TP/(TP+FP+FN)), Correction (TP/(TP+FP)), and Completeness (TP/(TP+FN)). Designating a polygon as a true positive assumed an IoU criterion greater than 50\% between predicted and reference polygons. Therefore, the present study used per-pixel metrics to evaluate the training and test steps using samples with dimensions of 128x128 pixels and per-polygon analysis to analyze the complete image after its reconstruction using the sliding window method.

\section{Results}
\subsection{Semantic Segmentation Metrics}
This investigation evaluated the performance of five deep-learning semantic segmentation architectures in different time-series compositions (Table 4). The first evident pattern is that the main metric (IoU) consistently decreases when reducing the number of images in the time series. One of the main reasons for this behavior is that, as the reflection of wind turbines varies in different images, using more images establishes a more consistent pattern, improving understanding of the deep-learning model. Furthermore, the time-invariant behavior of wind farms allows us to distinguish them from other mobile features such as ships.

Overall, LinkNet presented the best results, slightly better than U-Net++ and U-Net. These models not only offer a more straightforward path to restore images to their original size, but their architectures also ensure the retention of high-resolution features during upsampling. This characteristic makes them more suitable for tasks requiring detailed semantic segmentation, as with consistently small, uniformly scaled objects. Contrarily, the FPN (Feature Pyramid Network) and DeepLabv3+ models yielded significantly poorer results, potentially due to the inherent characteristics of the object under study, which were consistently small. DeepLabv3+ typically works best when applied to datasets featuring a range of scales and sizes. Likewise, the FPN model has a mechanism that leverages lateral connections between diverse feature maps, a less beneficial approach when dealing with uniformly small, unvarying objects.

Then, an effectiveness analysis of the proposed augmentation strategy (temporal shuffling) took place by comparing the results with and without its application for the best-training model (LinkNet) and the time series with 5, 10, and 15 images. Table 5 indicates that the results for the different time series obtained without applying the proposed augmentation strategy show a significant loss of accuracy, reaching around 18\% of the F-score and 25\% of IoU.

The time series with 15 images exhibited a lower performance, resulting in a 24.45\% decrease in IoU compared to the trained model with the proposed augmentation. Similarly, the time series with ten images showed a decline of 26.2\%, while the time series with five images experienced a 23.9\% decrease in performance. These findings highlight the benefits of our augmentation strategy and its ability to enhance the model's performance across various time series configurations.

Figure 3 provides a visual demonstration of the predictive capabilities of the LinkNet model when subjected to various time-series configurations. The observed variation within the predictions is notably minimal when utilizing the proposed augmentation strategy. Despite the significant IoU difference, the models identify the targeted objects proficiently. The leading cause of this robust performance is the small size of the objects in consideration. Even a minuscule shift in a pixel or two can instigate a considerable impact on the overall performance metrics, demonstrating the fine-tuned sensitivity of the models.

\begin{figure}[!h]
	\centering %
	\scriptsize %
	\includegraphics[width=0.8\columnwidth]{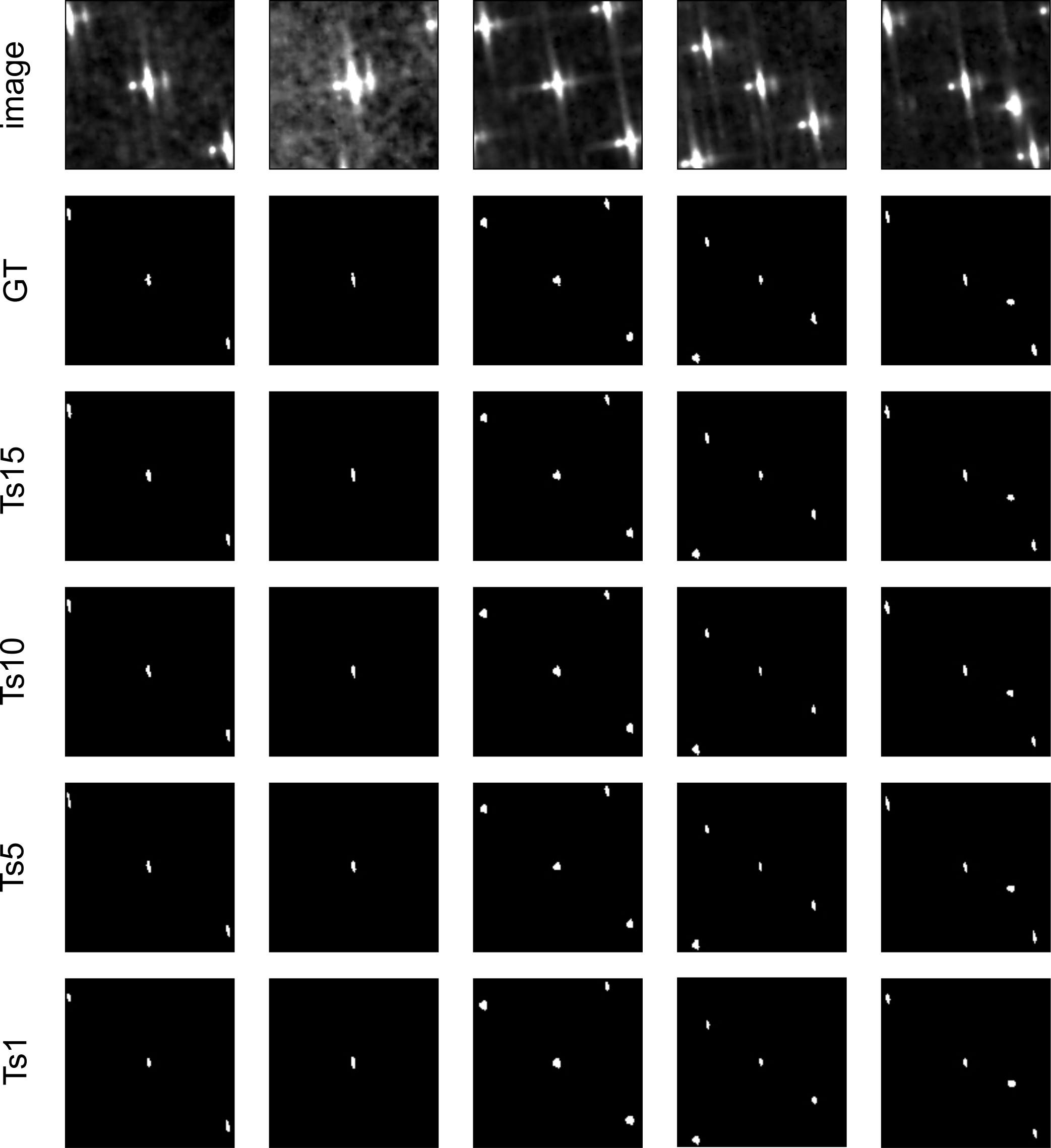}
	\caption{Original SAR images (images), ground truth (GT), and time series predictions with the LinkNet model using 15 (TS 15), 10 (TS 10), 5 (TS 5), and 1 (TS 1) images. * Denotes the time series predictions without the proposed augmentation strategy.}
	\label{fig:fig3}
\end{figure}

Notably, deploying more images led to remarkably consistent predictions and demonstrated a reduced propensity for misclassification of other high-backscatter targets. Conversely, a marked decrease in the model's ability to discern certain features occurred when the augmentation strategy was not applied. This result indicates the model's tendency to memorize patterns within the time series that do not find representation within the test set. Therefore, the proposed augmentation strategy offers an ability to improve the predictive ability and resilience of the model in maintaining object recognition, providing perspectives on future reliable predictions of other targets.

\begin{table}[!h]
\centering
\setlength\extrarowheight{-4pt}
\caption{Semantic segmentation results considering the overall accuracy (OA), precision, recall, f-score, and intersection over union (IoU) metrics for the time series with 15, 10, 5, and 1 image.}
\begin{tabular}{l|l|l|l|l|l}
\hline
Model & OA & Precision & Recall & F-score & IoU \\
\hline
\multicolumn{6}{c}{Time Series with 15 images} \\
\hline
LinkNet    & 99.96 & 87.62 & 92.29 & 89.90 & 81.65 \\
U-Net      & 99.96 & 87.75 & 91.59 & 89.63 & 81.21 \\
U-Net++    & 99.96 & 87.43 & 92.10 & 89.71 & 81.33 \\
DeepLabv3+ & 99.93 & 79.82 & 86.72 & 83.13 & 71.13 \\
FPN        & 99.92 & 79.81 & 85.01 & 82.33 & 69.96 \\
\hline
\multicolumn{6}{c}{Time Series with 10 images} \\
\hline
LinkNet    & 99.95 & 86.65 & 89.11 & 87.86 & 78.35 \\
U-Net      & 99.95 & 86.04 & 88.55 & 87.28 & 77.42 \\
U-Net++    & 99.94 & 84.83 & 89.44 & 87.07 & 77.11 \\
DeepLabv3+ & 99.92 & 78.82 & 83.60 & 81.14 & 68.26 \\
FPN        & 99.91 & 75.76 & 85.47 & 80.32 & 67.11 \\
\hline
\multicolumn{6}{c}{Time Series with 5 images} \\
\hline
LinkNet    & 99.93 & 82.94 & 86.03 & 84.45 & 73.09 \\
U-Net      & 99.93 & 83.29 & 85.54 & 84.40 & 73.01 \\
U-Net++    & 99.93 & 83.06 & 86.47 & 84.73 & 73.51 \\
DeepLabv3+ & 99.90 & 73.76 & 82.88 & 78.06 & 64.01 \\
FPN        & 99.90 & 72.94 & 83.00 & 77.64 & 63.46 \\
\hline
\multicolumn{6}{c}{Single Image} \\
\hline
LinkNet    & 99.90 & 76.63 & 78.42 & 77.52 & 63.29 \\
U-Net      & 99.90 & 71.29 & 85.51 & 77.76 & 63.61 \\
U-Net++    & 99.90 & 75.34 & 79.08 & 77.16 & 62.82 \\
DeepLabv3+ & 99.88 & 67.70 & 80.38 & 73.50 & 58.10 \\
FPN        & 99.88 & 69.76 & 77.96 & 73.63 & 58.27 \\
\hline
\hline
\end{tabular}
\label{tab:tab4}
\end{table}

\begin{table}[!h]
\centering
\setlength\extrarowheight{-4pt}
\caption{Semantic segmentation results using the LinkNet model for time series with 15, 10, and 5 images without the proposed augmentation strategy. “OA” is overall accuracy, and “IoU” is Intersection over Union.}
\begin{tabular}{l|l|l|l|l|l}
\hline
Time Series & OA & Precision & Recall & F-score & IoU \\
\hline
15   & 99.90 & 90.89 & 60.67 & 72.77 & 57.20 \\
10   & 99.86 & 63.87 & 73.96 & 68.55 & 52.15 \\
5    & 99.89 & 87.18 & 53.39 & 66.22 & 49.50 \\
\hline
\end{tabular}
\label{tab:tab5}
\end{table}

\subsection{Full image classification}

The full image reconstruction of the semantic segmentation used the sliding window approach. The analysis of the per-object metrics considered the region with the highest concentration of offshore wind farms and the different time series composed of 1, 5, 10, and 15 images (Table 6). The results indicated a significant initial improvement with increasing images from 1 to 5, decreasing the number of false positives from 14 to 2 objects, and nullifying the presence of 2 false negatives. Among the two remaining false positives with five images, the increase to 10 or 15 images eliminated one. Therefore, the subsequent transitions from 5 to 10 and 15 images show minor improvements after reaching a certain threshold (5 images in this case).

\begin{table}[h!]
\begin{tabular}{l|l|l|l|l}
\hline
\multirow{2}{*}{} & \multicolumn{4}{l}{Number of images used in the time series} \\
\hline
                  & 1                               & 5                              & 10                             & 15                              \\
                  \hline
TP                & 296                             & 297                            & 297                            & 297                            \\

FP                & 14                              & 2                              & 1                              & 1                              \\

FN                & 1                               & 0                              & 0                              & 0                              \\
\hline
Overall Quality   & $\frac{296}{296 + 14 + 1} \approx 0.954$ & $\frac{297}{297 + 2 + 0} \approx 0.993$ & $\frac{297}{297 + 1 + 0} \approx 0.997$ & $\frac{297}{297 + 1 + 0} \approx 0.997$ \\

Correctness       & $\frac{296}{296 + 14} \approx 0.955$     & $\frac{297}{297 + 2} \approx 0.993$     & $\frac{297}{297 + 1} \approx 0.997$     & $\frac{297}{297 + 1} \approx 0.997$     \\

Completeness      & $\frac{296}{296 + 1} \approx 0.997$      & $\frac{297}{297 + 0} = 1$               & $\frac{297}{297 + 0} = 1$               & $\frac{297}{297 + 0} = 1$    \\           
\hline
\end{tabular}
\caption{Metrics and Quality Measures for Time Series Images}
\label{table:time_series_metrics}
\end{table}

The raster-to-polygon conversion of the semantic segmentation results in Geographic Information Systems (GIS) proved highly efficient in representing the offshore wind farm at the instance level. Figure 4 shows the results in polygon vectors of offshore wind plants obtained by the Linknet model considering a 15-image time series. The isolated behavior of wind farms, as opposed to other overlay or contact targets such as cars \citep{mou2018vehicle}, provides a semantic-to-instance segmentation conversion facility.

\begin{figure}[!h]
	\centering %
	\scriptsize %
	\includegraphics[width=0.8\columnwidth]{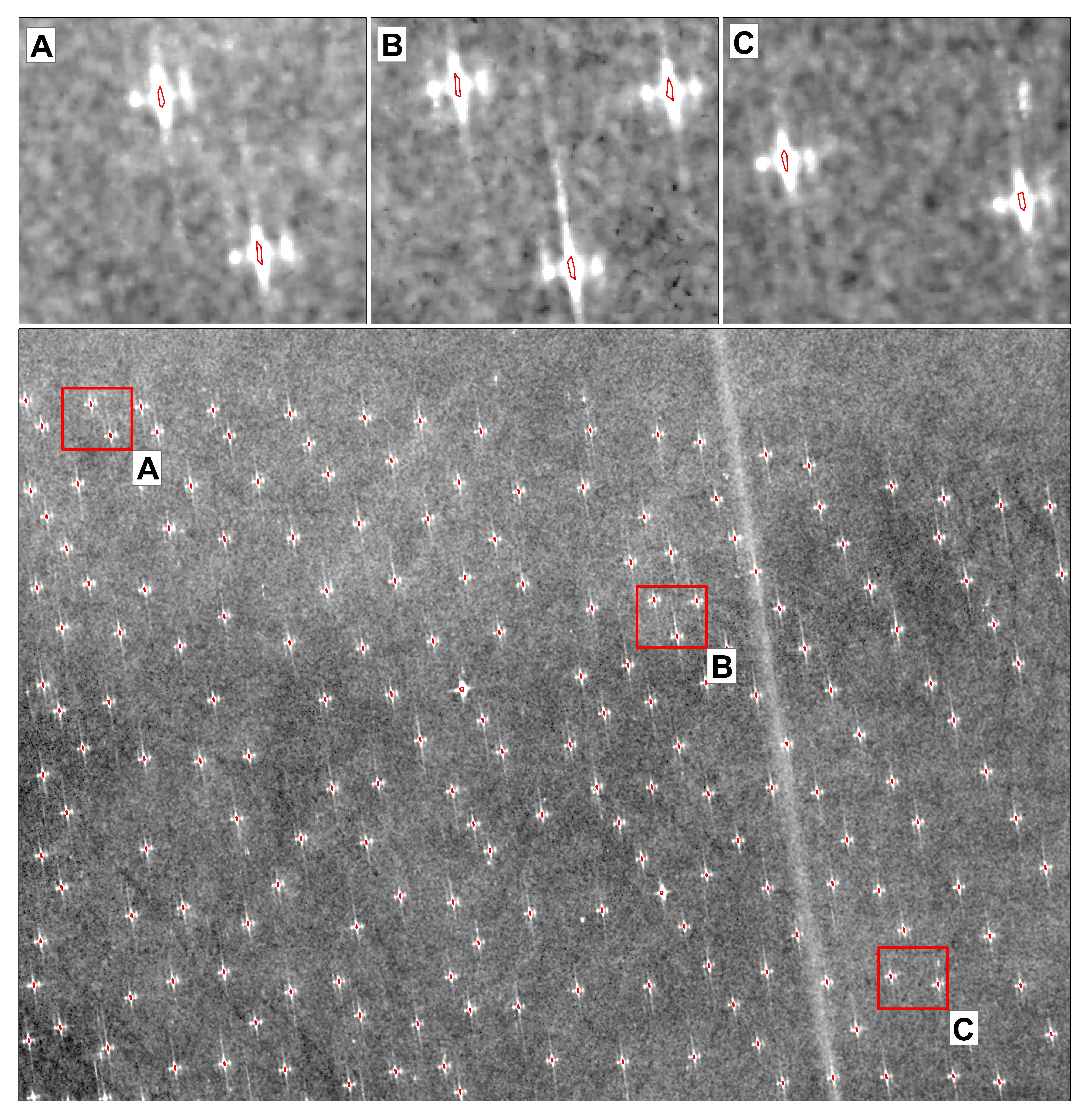}
	\caption{Semantic segmentation results for a large image using the sliding window approach for a region with a high concentration of offshore wind farms. Red polygons derived through the raster-to-polygon conversion of the LinkNet prediction superimposed on Sentinel-1 VV polarization image delineated the offshore wind plant features.}
	\label{fig:fig4}
\end{figure}

Figure 5 demonstrates the errors from using a single image in predicting offshore wind plants marked in blue. Although the model demonstrated competence in recognizing brightness and identifying most targets, it also generated many false positives regarding the other models using time series.

\begin{figure}[!h]
	\centering %
	\scriptsize %
	\includegraphics[width=0.9\columnwidth]{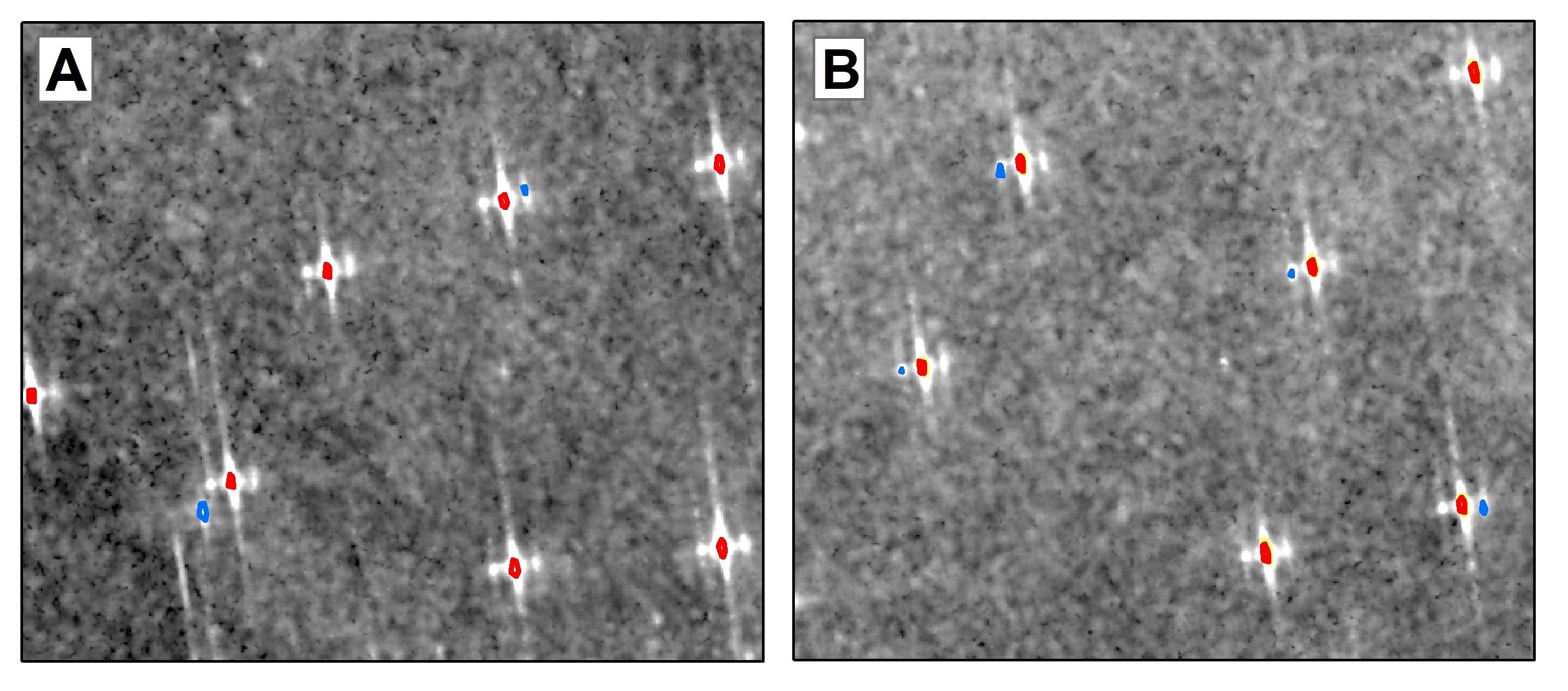}
	\caption{Results using a single image from the time series and their corresponding errors. Correct results are in red, and incorrect classifications are in blue.}
	\label{fig:fig5}
\end{figure}

\pagebreak
\section{Discussion}
This research sought to develop a new approach using time series data with different CNN models to detect offshore wind farms. The developed strategy comprised generating a data set with different amounts of images and an innovative augmentation procedure that promotes the shuffling of the chronological sequence, which allows learning with different sequential contexts.

\subsection{Dataset Importance}
Anticipating a significant increase in dependence on renewable energies in the coming years and offshore wind farm growth worldwide, the development of automatic inspection methods is critical. Understanding regulatory and environmental constraints and spatial patterns will be crucial to managing this projected growth effectively. Considering these challenges, the offshore wind farm dataset developed innovates by being the first in deep learning semantic segmentation and having multi-channels referring to time series images, differentiating itself from other databases for object detection and using only a channel. The proposed dataset considered ten regions of the United Kingdom (a global leader in offshore wind energy), containing more than five thousand patches. The dataset also includes different time-series components, containing up to fifteen images, ensuring a broad scope of scenarios, backscatter patterns, mobile targets, and noise. Despite its UK-centric composition, the dataset can be applied to other locations and, if necessary, adjusted by entering new data based on our guidelines.

\subsection{Time Series augmentation effect}
This research employed an innovative augmentation strategy explicitly designed for semantic segmentation models of invariant targets, involving a randomized rearrange of image positions within a time series and offering a unique and efficacious way to mitigate biases associated with ordering images within the temporal sequence. Thus, sequential backscatter variations (at a given chronological position) in the time series are shuffled in the samples, allowing the adaptation of the training model for the invariant-target segmentation in other unknown Sentinel-1 images. The proposed augmentation strategy also presents a promising approach to detect other invariant targets using radar imagery, such as buildings and fixed infrastructure. However, this strategy may not be universally beneficial when the chronological order of the frames is significant due to the disruption of temporal ordering, such as in studies based on phenology. 

The comparative analysis of the interval dimension of the time series demonstrated that the data incorporation promoted benefits. Inserting target backscatter variations helps the algorithm navigate several degrees of complexity, promoting a deeper understanding of the intricate relationships. The broader time series eliminates noise and mobile elements and reiterates the mapped features. For example, a ship can exhibit similar patterns and confusion when analyzing a single image. However, the model improves perception and accuracy as the time series increases. Nevertheless, increasing the time series without limits can also lead to complications. For example, extending the time series too far into the past may include a period when the wind farm was not installed, negatively affecting model performance.

\subsection{Segmentation model analysis}

The accuracy comparison among the semantic segmentation models for offshore wind farm detection showed a group composed of LinkNet, U-Net, and U-Net++ that achieved better performance than a second group, composed of DLv3+ and FPN, with lower precision metrics. The need for more investigation hampers the comparison with other deep-segmentation studies for detecting offshore wind farms. Among the few studies, \cite{xu2022dynamic} performed a segmentation based on machine learning methods for offshore wind farms with disadvantages compared to deep-learning approaches widely described in the literature, particularly the enhanced understanding of context and the scalability offered by CNNs. 

Although this study focused on offshore wind farms with SAR images, other deep semantic segmentation investigations have been developed on onshore wind farms using high-resolution optical images \citep{de2023data, han2018targets, manso2020optimizing, manso2021first, schulz2021deteektor}. Despite the differences, the results between the models for both situations, using optical and radar images, show apparent similarities. The results by \cite{de2023data} that detect onshore wind farms using CBERS optical images present the same architecture results based on IoU and F-score, evidencing the LinkNet architecture with Eff-B7 backbone as the best model, followed by the U-Net and U-Net++ models, while DLv3+ and FPN showed more than a 3\% difference from the three best models. \cite{manso2020optimizing}, comparing LinkNet and U-Net using aerial images, also observed better performance of LinkNet. The other onshore wind farm studies did not make comparisons, using only one architecture. A possible explanation for the difference between the models is that the LinkNet, U-Net, and U-Net++ methods present better results in capturing fine details and preserving spatial information, which becomes suitable for small objects with fixed dimensions. On the other hand, DLv3+ and FPN methods offer better performance for targets with size and shape variations. 

Furthermore, the research demonstrated a simple procedure for semantic-to-instance segmentation conversion using GIS, facilitated by the absence of overlapping objects. The occurrence of having only isolated targets avoids the need to use other procedures when the targets present agglutination, such as edges to isolate the interior of objects and multitask learning \citep{mou2018vehicle} or multiclass learning \citep{de2022bounding, de2022rethinking}. The dataset complies with the COCO annotation guidelines using a specific program that automatically converts GIS data to the COCO annotation format, allowing the use of advanced Mask-RCNN models \citep{de2022panoptic, carvalho2020instance}.

\section{Conclusion}
This research presented a pioneering application of deep segmentation for offshore wind farms based on the Sentinel-1 time series, proposed a novel augmentation strategy, compared five CNN architectures, and used GIS to obtain an instance-level segmentation of these features. Sentinel-1 radar images revealed a valuable contrast for detecting offshore wind farms. The offshore wind plant dataset elaborated in this study is not limited to a single date (single channel), using a time series (multi-channel) that, with the application of the augmentation strategy, presents a chronological sequence scrambling to increase the detection potential of invariant targets in unknown images. Time series instead of a single image favors eliminating other mobile high-backscatter features (e.g., ships) and reinforcing differences with other fixed objects (e.g., oil platforms). In all semantic segmentation models, the IoU and F-score metrics consistently increase as the number of images in the time series increases. The LinkNet achieved the best results, slightly surpassing U-Net++ and U-Net, while the FPN and DeepLabv3+ models yielded the worst results. Using the novel augmentation strategy represented a significant improvement in accuracy, increasing by around 18\% of the F-score and 25\% of the IoU in the different time series with 5, 10, and 15 images. These results demonstrate that the proposed augmentation strategy is a promising method to minimize biases in detecting invariant targets. Per-object accuracy metrics allow the evaluation of a balance in determining the number of images in the time series and improving performance to optimize the model, reaching 5 or 10 images in this investigation. The semantic-to-instance segmentation conversion is highly effective, as the offshore wind plants are isolated without contact, promoting advantages such as simplified training data and accurate semantic segmentation methods. Therefore, as the world's dependence on renewable energy grows, this research proves the potential of deep learning in detecting offshore wind farms, which is essential to improve the management and sustainability of natural resources.

\section*{Declaration of competing interest}
The authors declare that they have no known competing financial interests or personal relationships that could have appeared to influence the work reported in this paper.

\section*{Data Availability Statement}
The dataset and code can be obtained by contacting the corresponding author upon reasonable request.

\section*{Acknowledgments}
The authors are grateful for financial support from the CNPq fellowship (Osmar Abílio de Carvalho Júnior). This study was partly financed by the Coordination for the Improvement of Higher Education Personnel (CAPES) – Finance Code 001. Finally, the authors acknowledge the contribution of anonymous reviewers.

%Bibliography
%\clearpage %Put references in their own page
\bibliographystyle{elsevier-model5-names}\biboptions{authoryear}
\bibliography{bibliography.bib}

\end{document}